\documentclass[letterpaper, 10 pt, conference]{ieeeconf}

\usepackage{epsfig} 
\usepackage{amsmath}
\usepackage{float}
\usepackage[utf8]{inputenc}
\usepackage{graphicx}
\usepackage{multicol}
\usepackage{footmisc}
\usepackage{amssymb}
\usepackage{amsmath}
\usepackage{amsfonts}
\usepackage{algorithm}
\usepackage[noend]{algorithmic}
\usepackage{tikz}
\usepackage{hyperref}
\usetikzlibrary{math,shapes}
\usepackage[caption=false, font=footnotesize]{subfig}
\usepackage{listings}
\usepackage{xcolor}

\IEEEoverridecommandlockouts
\title{\LARGE \bf Rocket Landing Control with Grid Fins and Path-following using MPC}
\author{Junhao Yu, Jiarun (Sammy) Wei$^1$%
\thanks{$^1$Couse Project for ME 231A.
{\tt\small \{junhaoyu, sammywei\} @berkeley.edu}}}

\begin{document}

\maketitle    
\thispagestyle{empty}
\pagestyle{empty}
\begin{abstract}
 This report is a summary of University of California, Berkeley's ME C231A final project. In this project, we attempt to optimize a landing trajectory of a rocket. The goal is to minimize the total fuel consumption during the landing process using different techniques. Once the optimal and feasible trajectory is generated using batch approach, we attempt to follow the path using a Model Predictive Control (MPC) based algorithm, called Trajectory Optimizing Path-following Estimation from Demonstration (TOPED), in order to generalize to similar initial states and models, where we introduce a novel cost function for the MPC to solve. We further show that TOPED can follow a demonstration trajectory well in practice under model mismatch and different initial states.

\end{abstract}


\section{Problem Statement}
Given a rocket dynamic system $f$ with states $m,x,y,\dot{x},\dot{y},\theta, \dot{\theta}$ and inputs $F, \delta$: $m$ is total mass of the rocket, $x$ is horizontal position of the rocket, $y$ is vertical position of the rocket,
$\dot{x}$ is horizontal velocity of the rocket, $\dot{y}$ is the vertical velocity of the rocket, $\theta$ is the angle between the center line of the rocket and the gravity line,  $\dot{\theta}$ the rate of change of $\theta$, $F$ is magnitude of the thrust force, $\delta$ is the angle between the thrust force and the center line of the rocket, $K$ is the ratio of $F$ and the rate of mass change $\dot{m}$, $J$ is moment of inertia of the rocket, and $l$ is length of the rocket, we wish to minimize the cost of fuel during the landing process of the rocket given the number of time steps. This objective can be formulated as an optimization problem, and solved using batch method. Once the optimization is solved, we obtain a planned optimal trajectory $\tau^*$ that realizes our objective. Then we plan to use an MPC controller $\mathcal{C}$ to follow this planned trajectory as closely as possible. We define a cost function for MPC which is a weighted sum of the original cost function, L2 norm of states mismatch, L2 norm of input mismatch, and slack variables.

According to the above configuration, the state space representation and dynamic model of the system are as follows:
$$
    z(k)=\begin{bmatrix}
    m(k)&
    x(k)&
    y(k)&
    \dot{x}(k)&
    \dot{y}(k)&
    \theta(k)&
    \dot{\theta}(k)
    \end{bmatrix}^T
$$
The system dynamics for the \textbf{original system without the grid fins} are defined as:
$$
    z(k+1)=z(k)+T_s\cdot\begin{bmatrix}
     -\frac{F}{K}\\
     \dot{x}(k)\\
     \dot{y}(k)\\
    -\frac{F\sin(\theta+\delta)}{m}\\
    g-\frac{F\cos(\theta+\delta)}{m}\\
    \dot{\theta}(k)\\
     -\frac{Fl\sin(\delta)}{(2J)}
    \end{bmatrix}
$$
with the control input: $
    u=\left[F\; \delta\right]^T
$.

The system dynamics for the \textbf{new system with grid fins} are defined as:
$$
    z(k+1)=z(k)+T_s\cdot\begin{bmatrix}
     -\frac{F}{K}\\
     \dot{x}(k)\\
     \dot{y}(k)\\
    -\frac{F\cos(\theta+\delta)}{m}-\frac{\gamma\rho A_g \dot{x}(k)^2}{2m}\\
    g-\frac{F\cos(\theta+\delta)}{m}-\frac{\gamma\rho A_g \dot{y}(k)^2}{2m}\\
    \dot{\theta}(k)\\
     -\frac{Fl\sin(\delta)}{(2J)}
    \end{bmatrix}
$$
with control input: $
    u=\left[F\; \delta \; \gamma \right]^T
$, where constant $A_g$ denotes the maximum area of the grid fins and $\gamma$ is a grid fins angle coefficient such that $0.8\leq\gamma\leq1$.

\section{Related Work}
Our project is highly related to ideas about optimal control and path following techniques and learning from demonstration \cite{zhang2016health, zhang2021robots, zhang2020dex, elmquist2022art, sim2019personalization}. In \cite{pascucci2015model}, the authors demonstrated an MPC-based approach to follow a precomputed trajectory (path) in the setting of rocket landing guidance. Generally, learning from demonstration is beneficial in scenarios when the demo model is similar to the actual model of the system \cite{pan2022tax, pan2023tax, eisner2022flowbot3d, zhang2023flowbot++}. Therefore, the agent can learn from demo in order to generalize to slightly different models and initial states. Recent literature has shown a variety of methods to learn from demonstration \cite{shen2024diffclip, yao2023apla}. For example Dataset Aggregation (DAgger) method \cite{ross2011reduction, devgon2020orienting, avigal20206, avigal2021avplug, lim2021planar, lim2022real2sim2real} requires human labeling of a dataset and aggregates the labeled dataset to the agent training data in order to converge to optimal trajectory. Given a feasible trajectory, the agent can follow the trajectory in a safe manner by simultaneously learning the safe set of MPC, as proposed in \cite{thananjeyan2019extending}.

\section{Methods and Experiments}
\subsection{Optimal Landing Trajectory with Grid Fins}
Consider the above dynamics models, the original system only has thrust vector control while the new system has both TVC and grid fins. By adding grid fins, the system gained extra lift and torque and thus outperformed original system with merely Thrust Vector Control. We then solve for optimal trajectory using fmincon. Based on our prediction, the rocket with grid fins would have different trajectory due to the extra lift and torque, resulting in both less fuel burned and better altitude angle ($\theta$) control.

\subsection{MPC Path-following}
We propose a path-following algorithm, Trajectory Optimizaing Path-following Estimation from Demonstrations (TOPED). Given a demonstration trajectory that is feasible and optimal, we want to follow the trajectory $\tau^*$ using MPC even under slightly imperfect and uncertain model and different initial states. In TOPED, we require three different weight vectors, each representing how important a state or input is in our path-following setting. The weights can therefore be tuned to follow different states of interest. We also require the system dynamics $f'$ as the constraints in MPC's receding horizon optimal control problem. The optional slack variables $\lambda$ with weights $\epsilon$ are used when the receding horizon control problem is infeasible. In such cases, we add slackness to inputs, and solve the MPC again. Inspired by \cite{pascucci2015model}, we define cost function at time step $k$ as $Q(x_k, u_k)=\lVert W_x^T(x_k-x^*_k) \rVert _2^2 + \lVert W_u^T(u_k-u^*_k) \rVert _2^2 + W_JJ(x,u)^2$. The receding horizon optimal control problem that Algorithm \ref{alg:the_alg} is trying to solve is defined as:

\begin{equation}
\begin{aligned}
\min\;{\sum_{k=0}^{N-1}Q(x_k, u_k)+\epsilon^T\lambda}\\
\textrm{s.t. }\;x(k+1)=f'(x(k),u(k))\; \forall{k}\\
    x_k\in \mathcal{X},u_k\in \mathcal{U}\; \forall{k}\\
    x_0=x(0)\\
    \end{aligned}
\end{equation}
where $x^*$ and $u^*$ are reference states and actions from demo. System dynamics model $f'$ could potentially be different than the original model $f$ used in generating the optimal path by solving the optimal control problem.
\makeatletter
\newcommand\fs@betterruled{%
  \def\@fs@cfont{\bfseries}\let\@fs@capt\floatc@ruled
  \def\@fs@pre{\vspace*{5pt}\hrule height.8pt depth0pt \kern2pt}%
  \def\@fs@post{\kern2pt\hrule\relax}%
  \def\@fs@mid{\kern2pt\hrule\kern2pt}%
  \let\@fs@iftopcapt\iftrue}
\floatstyle{betterruled}
\restylefloat{algorithm}
\makeatother

\begin{algorithm}[t!]
\caption{Trajectory Optimizing Path-following Estimation from Demonstrations (TOPED)}
\begin{algorithmic}[1]
\label{alg:the_alg}
\REQUIRE Feasible demonstration path $\tau^*$; system model $f$; cost function $J$; simulation horizon $M$; weight factors $W_x$, $W_u$, $W_J$; MPC prediction horizon $N$, slack variables $\lambda$; slack variable weights $\epsilon$

\FOR{$t$ $\in \{1, ..., M\}$}
    \STATE Extract states and inputs from demo trajectory $\tau^*$ as $x^*$ and $u^*$
    \STATE Define loss function
    $Q(x,u) = \lVert W_x^T(x_t - x_t^*) \rVert_2^2 + \lVert W_u^T(u_t - u_t^*) \rVert_2^2 + W_JJ(x, u)^2 + \epsilon^T \lambda$
    \STATE Run MPC with prediction horizon $N$ to minimize $Q(x,u)$, initially without slack variables and their weights, $\epsilon^T \lambda = 0$, obtain $x^{opt}_t$ and $u^{opt}_t$, using $f$ as dynamics constraint
    
    \IF {INFEASIBLE}
        \STATE Add slackness to input constraints, and increase weights of slack variables as defined in $Q(x,u)$ and solve again until feasible
    \ENDIF
    \STATE $x_{opt} \gets x^{opt}_t$, $u_{opt} \gets u^{opt}_t$
\ENDFOR
\RETURN optimal path-following trajectory $x_{opt}, u_{opt}$ as $\tau^{return}$
\end{algorithmic}
\end{algorithm}

Given a precomputed (demonstration) optimal landing trajectory $\tau^*$, we use an MPC-based Algorithm \ref{alg:the_alg} to follow the trajectory as closely as possible. We introduce a Gaussian noise $\mathcal{N}$ to our model in order to add uncertainty to our model to make path-following more challenging. Specifically, we add noise to some of the states and inputs readings, making the model less perfect. Moreover, we start from three different initial states: we now start with different amount of fuel and a location either farther or nearer than the provided trajectory's initial state. Since in our scenario, we care about the horizontal and vertical locations of the rocket the most, we assign more weight to those two states. Therefore, the second and the third entries in $W_x$ should be significantly larger than others. Intuitively, we assign more weight to the states and inputs we wish to follow, and assign less weight to the less essential states and inputs. In our setting, we want to follow states $x$ and $y$, inputs $F$ and $\delta$ as closely as possible, so the weight vector we used was $W_x = \left[1, 5, 5, 0.5, 1, 0.1, 0.5\right]$ and $W_u = \left[2,2,0.5\right]$. In practice, we made the weight assigned on the original cost function $W_J = 2$. In the MPC controller of TOPED, we used a prediction horizon $N = 4$, and our simulation length is $M = 20$ because the provided descent trajectory lands to origin in 20 steps. In practice, we also found that when the initial states are hugely different than the provided initial state (e.g. $x$ a lot greater or smaller than 165), the problem becomes infeasible no matter how much slackness we add to the input variables. Therefore, we ignore such cases and only apply feasible initial states because the idea of path-following is to generalize to similar situations. We also do not enforce the rocket to land exactly at the origin. Instead, we add slackness to the final location of the rocket so that it is within 15 meters from the origin.

\section{Results}
\subsection{Optimal Descent Path Planning}
We plot comparison of landing trajectory, $\theta$, thrust command and fuel-burned percentage between the original system and the system enhanced with grid fins. The grid fin command is plotted as well. The plots are shown in Fig 1. Though the two paths (a) have similar `S' shape, we can tell the difference through altitude angle (b), fuel-consumption curve (c) and thrust command (d): The system with grid fins stabilized $\theta$ faster without any oscillation while the original system experienced an oscillation. With the extra lift provided by grid fins, the new system tends to glide rather than landing directly, which saves around 8 percent of carried fuel. 

\newcommand{\Tk}[1]{\includegraphics[width=40mm, clip]{#1}}
\begin{figure}
    \centering
    \begin{tabular}{c@{\hspace{5pt}}c@{\hspace{5pt}}}
    \subfloat[$x$ and $y$ path ]{\Tk{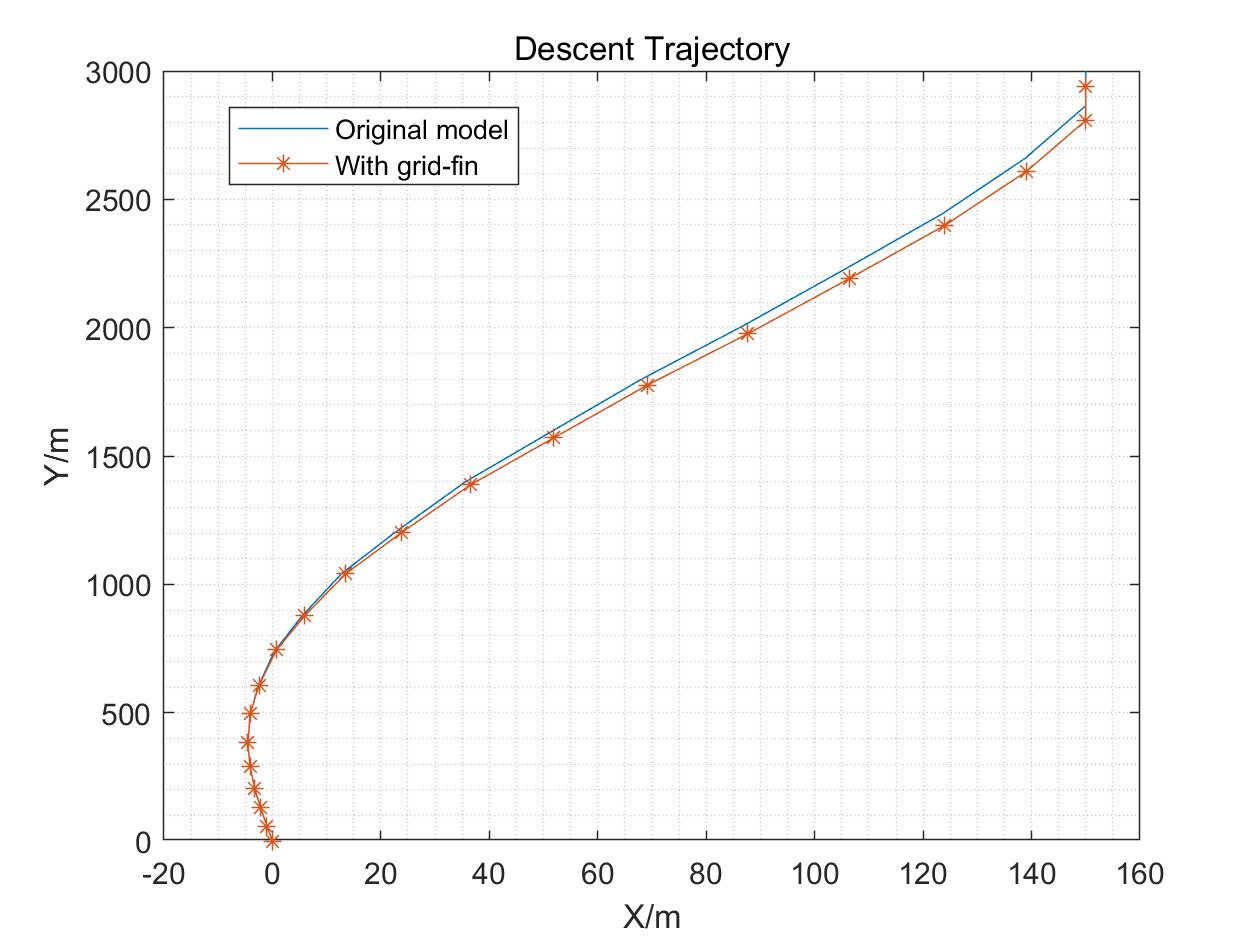}}&\subfloat[Altitude angle $\theta$ ]{\Tk{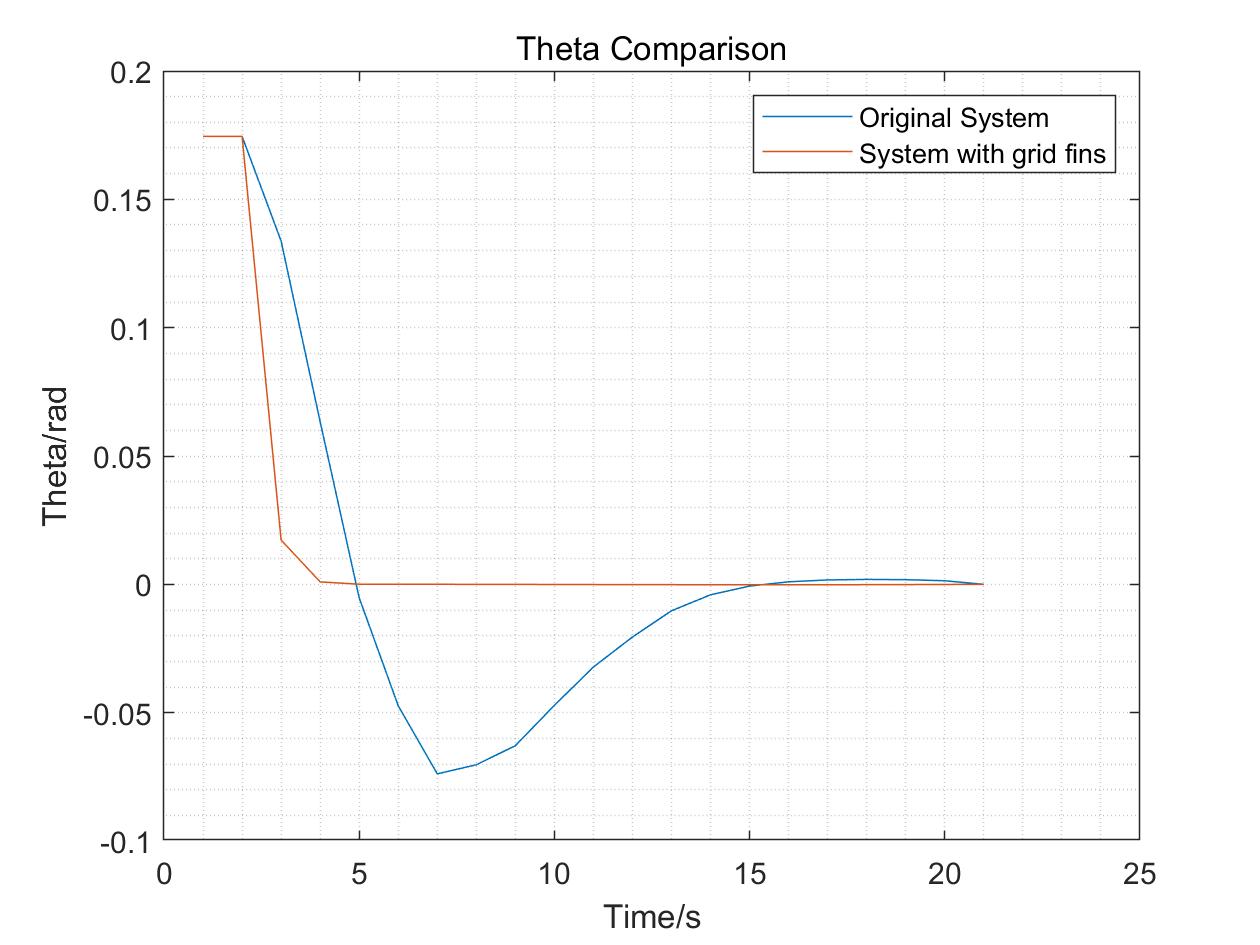}}\\
    \subfloat[Fuel Consumption Percentage]{\Tk{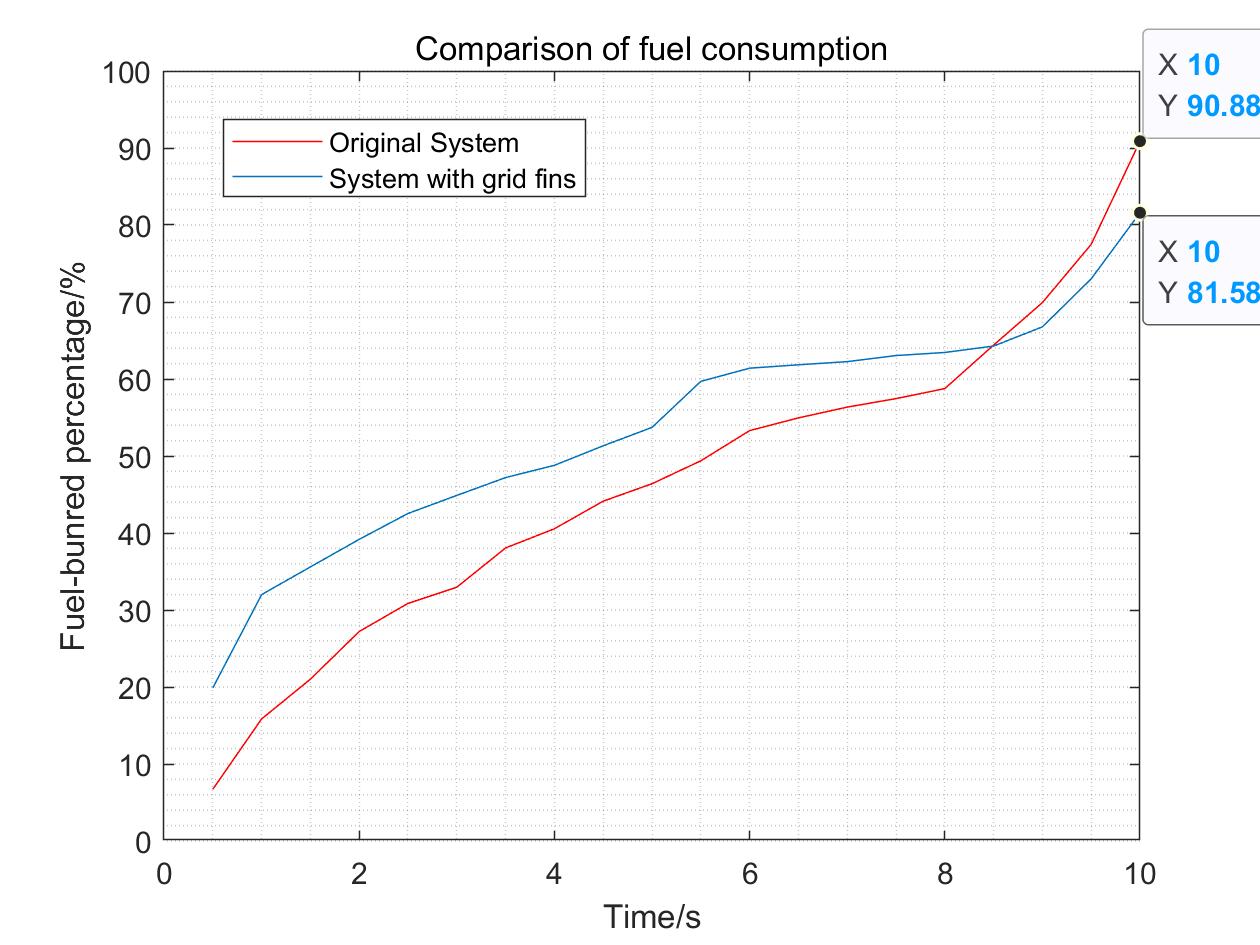}}&\subfloat[Thrust changes]{\Tk{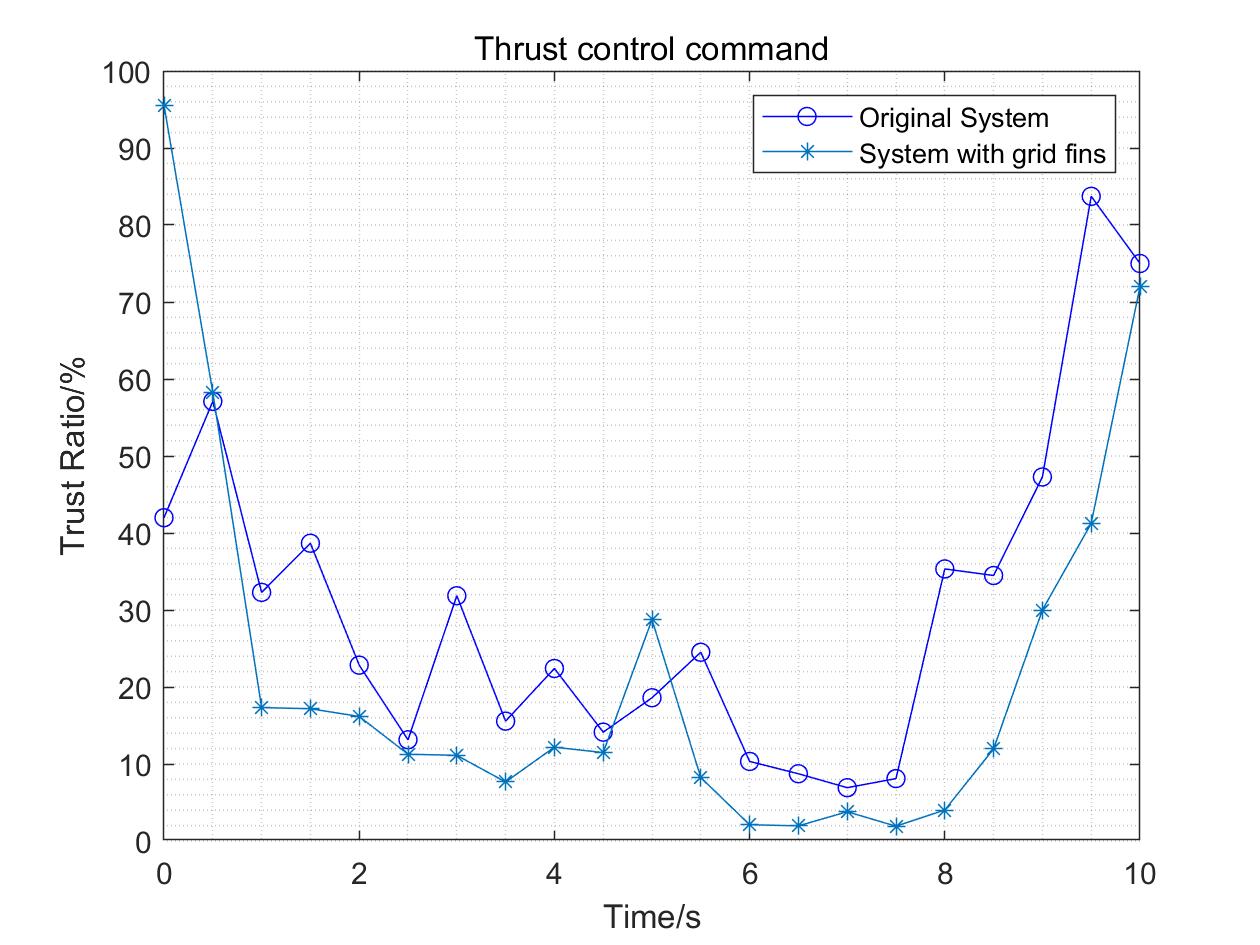}}\\
    \subfloat[Grid fin coefficient]{\Tk{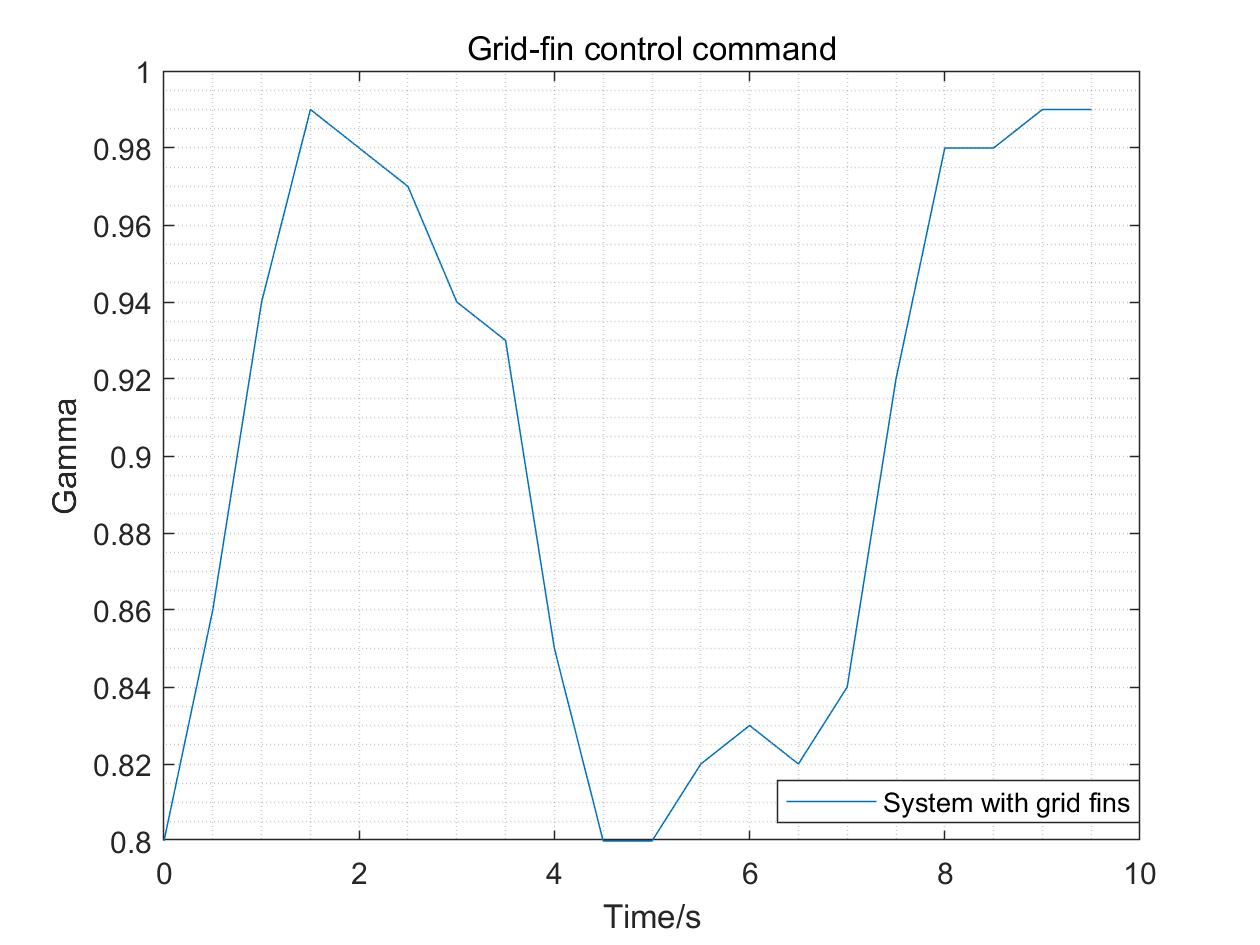}}
    \end{tabular}
    \caption{States and input changes in optimal landing trajectory for each model. \textbf{Top row:} $x$ and $y$ path during landing, and angle $\theta$ changes. \textbf{Middle row:} Fuel consumption changes and thrust changes. \textbf{Bottom row:} Grid fin coefficient changes.}
    \label{fig:my_label1}
\end{figure}

\subsection{MPC Path-following}
Our MPC path-following using Algorithm \ref{alg:the_alg} can follow the trajectory closely. To illustrate here, in Fig. 2, we plot out how TOPED follows the precomputed horizontal and vertical location $(x,y)$ of the rocket trajectory since our interest is to follow the descent path closely. The final locations under three different initial states of the rocket are within 10 meters from the origin. It can be observed that the MPC controller is trying to follow the demonstration path closely under slightly different models and initial states.
\begin{figure}
\includegraphics[width=9 cm]{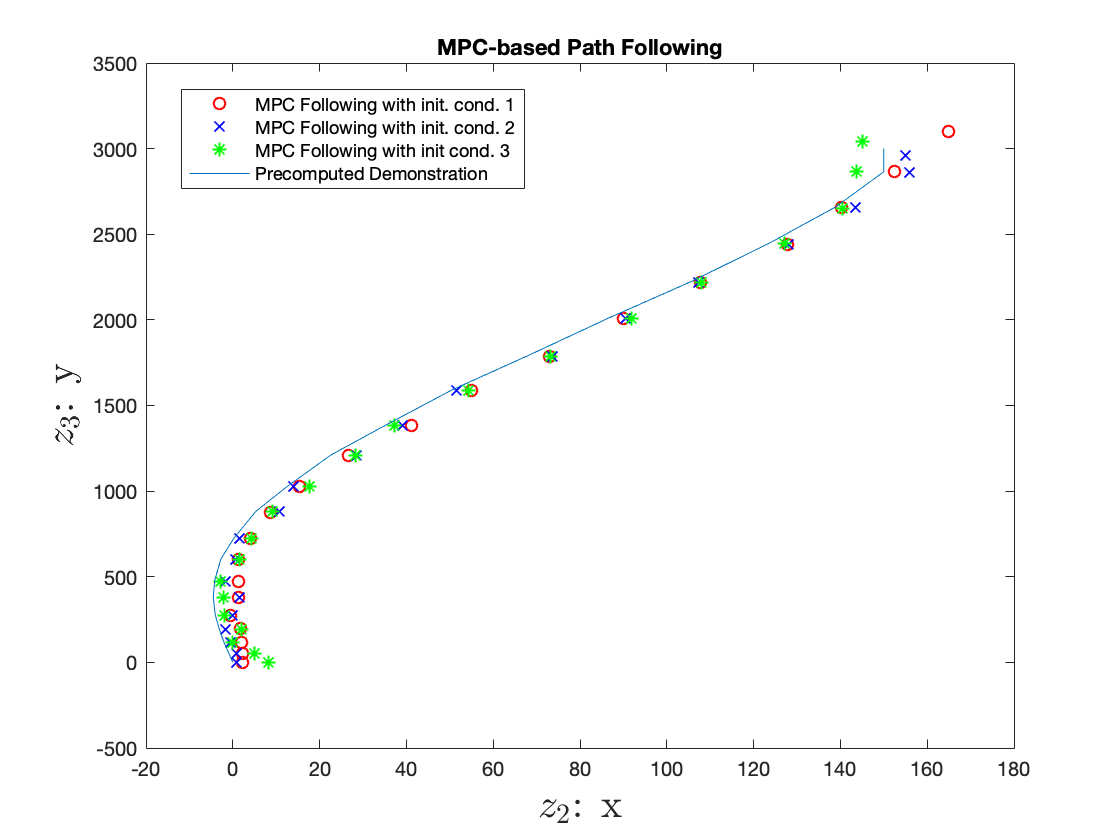}
\caption{Path following trajectory under model mismatch and different initial state. Initial condition 1 is 15 meters farther horizontally and 100 meters farther vertically, with 55kg less fuel to begin with. Initial condition 2 is 10 meters farther horizontally and 100 meters nearer vertically with 100kg less fuel to begin with. Initial condition 3 is 10 meters nearer horizontally and 100 meters farther vertically, with the same amount of fuel. All three test models are injected with random Gaussian noise $\mathcal{N}$.}

\end{figure}
One advantage of MPC is that in our experiment, the optimal reference demonstration path was generated in roughly 30 minutes on a PC with NVIDIA GTX1060. Since the MPC problem is shortsighted, unlike batch approach, we only look ahead 4 steps while closely following the demonstration trajectory, so the MPC path-following problem was solved in only about 56.75 seconds, which is significantly faster than batch approach, and the final trajectory is near optimal.
\section{Conclusion}
In this projects we explored some techniques used to optimize fuel consumption in rocket landing. Meanwhile, we proposed an MPC-based algorithm that can be used to follow a provided trajectory, and we experimentally showed that using Algorithm \ref{alg:the_alg}, the runtime complexity to solve the optimization problem decreases, while maintaining the near-optimality of the trajectory.

Since our system is non-linear, we do not attempt to make any convergence guarantee about TOPED, although we can say with similar initial states and models, Algorithm \ref{alg:the_alg} will drive the controller to follow the demonstration path. This notion of similarity can be quantitatively measured using KL-divergence if we introduce random variable noise to the model. Discussion of KL-divergence is beyond the scope of this report, but readers are welcomed to contact us for more information.

In terms of future work, we would explore our ideas in a scenario of three dimensional model, because our experiment was done in a 2D setting, which is a lot easier. Furthermore, since the optimal path was solved using optimization techniques, we would like to see if a human manipulated simulation dataset can be used to provide demonstration and reference. In this way, we no longer need to solve the optimization problem, and with such a dataset, one can simulate and optimize the landing and explore better trajectories using reinforcement learning techniques, and transfer to real environments, which can generalize to significantly more scenarios, not just limited to similar models, goals, and initial states.

\bibliographystyle{IEEEtran}
\bibliography{IEEEabrv,references}

\end{document}